# Spatiotemporal Pyramidal CNN with Depth-Wise Separable Convolution for Eye Blinking Detection in the Wild

Nguy Thi Lan Anh [1], Nguyen Gia Bach [2], Nguyen Thi Thanh Tu [1], Eiji Kamioka [2], and Phan Xuan Tan [2,*]

[1] School of Engineering Pedagogy, Hanoi University of Science and Technology, Hanoi, Vietnam;
Email: anh.ntl202019@sis.hust.edu.vn (N.T.L.A.), tu.nguyenthithanh@hust.edu.vn (N. T. T. T.)
[2] Graduate School of Engineering and Science, Shibaura Institute of Technology, Tokyo 135–8548, Japan;
Email: mg21501@shibaura-it.ac.jp (N. G. B.), kamioka@shibaura-it.ac.jp (E. K.)
*Correspondence: tanpx@shibaura-it.ac.jp (P.X.T)

*Abstract*—Eye blinking detection in the wild plays an essential role in deception detection, driving fatigue detection, *etc*. Despite the fact that numerous attempts have already been made, the majority of them have encountered difficulties, such as the derived eye images having different resolutions as the distance between the face and the camera changes; or the requirement of a lightweight detection model to obtain a short inference time in order to perform in real-time. In this research, two problems are addressed: how the eye blinking detection model can learn efficiently from different resolutions of eye pictures in diverse conditions; and how to reduce the size of the detection model for faster inference time. We propose to utilize upsampling and downsampling the input eye images to the same resolution as one potential solution for the first problem, then find out which interpolation method can result in the highest performance of the detection model. For the second problem, although a recent spatiotemporal convolutional neural network used for eye blinking detection has a strong capacity to extract both spatial and temporal characteristics, it remains having a high number of network parameters, leading to high inference time. Therefore, using Depth-wise Separable Convolution rather than conventional convolution layers inside each branch is considered in this paper as a feasible solution.

## I. INTRODUCTION

One of the critical signals that might reveal several aspects of human health and expression is the blinking of the eyes. And thus, a series of eye blinking patterns can be retrieved and used in several applications, including detecting driver drowsiness [1], face anti-spoofing [2], and communication between persons with disabilities [3], etc. In the driving or biometric-based surveillance scenarios, one important advantage of this approach is to receive cropped eye images of a person as inputs, and thus authority personnels only need to collect eye images data of different people, but not their full body, raising no concern in privacy or bias issues. In the education scenario, this method is encouraged to be used in online learning, in which the webcam can capture and analyze eye blinking patterns to detect the student's cognitive ability or mind wandering problems.

There has been a growing number of research in recent years to propose methods for eye blinking detection, ranging from the use of traditional signal processing, image processing, to deep learning. However, the majority of them don't take into account eye blinking in the wild. Most currently available eye-blink detection datasets along with the proposed methods carried out on those datasets are typically taken in confined indoor settings with partially inactive subjects. Nevertheless, eye-blink detection in the wild is more favored in some real-world application circumstances. For instance, eye-blink visual data may be covertly gathered using concealed cameras during the deception detection phase, in either unrestricted interior or outdoor situations. In this instance, performance must fundamentally be ensured via an efficient and real-time eyeblink detection technique in the field.

Additionally, when predicting eye blinks, there remain two issues. The first problem is that, depending on the situation, the distance between the face and the camera varies; and thus, the derived eye pictures may have different resolutions. As a result, a detection model may face a hurdle when trying to learn at various resolutions for the purpose of blinking detection. The second problem is that the detection model requires very low inference time in order to attain real-time performance. The inference time is how long it takes for a forward propagation from a given input sequence of eye images to a decision of blinking or non-blinking. Therefore, a lightweight detection model with a low number of parameters is needed for faster inference time.

Going further into the first issue, when detecting eye blinking in the wild, the distance between the subject's face and the camera may differ depending on, for example, the subject's standing position during face verification or recognition. As a result, the eye images may have low or high resolution and they must be resized into a constant size for the input of a deep learning -based detection model. When eye pictures are extracted in distant environment situations such as CCTV or house front-door cameras, the extracted eye images typically have poor quality and require upsampling to gain additional features for the model to learn. In contrast, when eyes are captured at close range (i.e., when looking at a computer screen, performing eye biometric verification, in-car camera monitoring driver etc.) much higher resolutions are obtained. However, when increasing the batch size during training to improve performance of deep learning -based detection models, high resolution images occupy a large amount of memory. And thus, sometimes researchers are forced to limit the batch size due to memory restrictions of hardware. One possible solution is downsampling of higher resolution images to train with larger batch size while having limited memory [7]. One question could be how upsampling and downsampling methods may affect the performance of classification models by modifying input features. Therefore, in this paper, we first analyze the effects of upsampling and downsampling interpolation techniques on the detection of eye blinking.

Going further into the second issue, real-time detection applications relating to images often rely on Convolutional Neural Network (CNN) -based models for its excellence in feature extraction [5], and researchers often look for a deeper, or more complex model architecture to gain improvement in its detection performance. This usually trades off with higher computation complexity by having a large number of network parameters, resulting in high inference time for new data and limiting real-time performance. Regarding eye blinking detection using CNN-based approach, a spatial-temporal CNN with a pyramidal bottleneck network [6] has recently demonstrated superiority over existing methods. However, their highest performance model of three pyramids with three branches (P3B3) remains having a high number of parameters (~ 7.6 million parameters), despite having great ability to extract both spatial and temporal features due to having more pyramids (deeper) and more branches (larger). Hence, it is imperative to decrease the number of network parameters while increasing model complexity. To this end, in this paper, we proposed to employ Depth-wise Separable Convolution layers, instead of traditional convolution layers inside each branch.

The remaining sections of the paper are organized as follows: Section II presents related studies about the evaluation of image resolution on deep learning performance and eye blinking detection methods. Section III describes the workflow of the eye blinking detection model, then focuses on the preprocessing step, and introduces a proposal for eye blinking detection step. Section IV presents the experiments and the obtained results. Section V discusses the obtained results. Finally, Section VI draws some conclusions and points towards future works.

## II. LITERATURE REVIEW

This section briefly reviews recent studies relating to 1) evaluation of interpolation methods and image resolutions on detection performance of a classification model, and 2) eye blinking detection methods from eye state estimations to deep learning based predictions and an in-the-wild eye blinking dataset.

Regarding analyzing the effects of interpolation and image sizes on the classification model, Bekhouche and Kajo *et al.* [6] employed picture downsampling to test the impact of five different pixel interpolation techniques (nearest neighbor, bilinear, Hamming window, bicubic, and Lanczos interpolation) on the prediction accuracy of a CNN for the aim of diagnosing on medical images. The five-pixel interpolation algorithms were obtained from Python Pillow's image-processing module to interpolate the image data for CNN. It was shown that selecting hamming or bicubic for downsampling may offer more accuracy when compared to other interpolation algorithms. However, the paper's explanation of how they outperform the other interpolation methods in the context of medical images is unclear.

In terms of analyzing suitable image resolutions, Sabottke and Spieler *et al.* [4] tracked CNN performance as a function of picture resolution for applications in diagnostic radiology on a dataset of chest radiographs from the National Institutes of Health. The performance of CNN was evaluated using the Area Under the receiver operating characteristic Curve (AUC) and label accuracy. For binary decision networks, maximum AUCs were obtained at picture resolutions between 256×256 and 448×448 pixels. Additionally, the biggest fractional increase in AUC was found to be obtained at higher picture resolutions (320×320 pixel) when compared with lower resolution (64×64 pixel) inputs.

Regarding the classifier for eye blinking detection, there has been extensive research on techniques ranging from using features tracker for eye state estimation, to implementing different deep learning architectures for a single or multiple eye image with an attempt to extract temporal features. Using a flock of KLT trackers, Drutarovsky and Fogelton *et al.* [10] introduced a motion-based eye blink detection system that tracks the initial ocular areas throughout time in order to increase the precision of the state estimation findings. On the Talking face dataset [16], the approach produced the best results across all metrics because it consists of 5000 frames taken from a video of a person engaged in conversation and this corresponds to about 200 seconds of recording. However, the ZJU dataset [17] yielded lower Recall because the Talking face contains a single European person, whereas the ZJU dataset contains Asian people whose eyebrows are mostly on average further from the eye. One-third of the method's failure to identify 71 eye blinkings on the ZJU was attributed to the Viola-Jones type algorithm's error. Because the video starts with a person with closed eyes, about 20 blinks are incomplete. Other failures generally happen as a result of extremely quick eye blinking that prevents the state machine from registering it.

Later investigations showed that the vast majority of currently used databases and algorithms in the field of eye blink detection are restricted to trials using a few hundred samples, as well as solitary sensors like face cameras. Daza and Morales *et al.* [8] proposed a novel multimodal database mEBAL with the support of sensors for eye blink detection and attention level estimation. They utilized three different sensors, an Electroencephalography (EEG) band to capture user cognitive activity and blinking events, a Near Infrared (NIR) and RGB cameras to record face expressions. The obtained outcomes have demonstrated the viability of using mEBL to train precise eye blink detectors under realistics acquisition settings. Nevertheless, their proposed VGG16-based model for eye blinking detection takes single images as input, and thus temporal features might not be captured. Additionally, the mEBL database was captured under a controlled environment, and thus their detection model might not generalize well in different scenarios.

To tackle the generalizability of a detection model, Hu *et al.* [9] proposed a novel dataset of eye blinking detection in the wild, HUST-LEBW [9], then suggested a modified LSTM architecture that can capture the multiscale temporal information of an eyeblink. The extraction of eye blinking features based on uniform LBP was then utilized. It simultaneously records both motion and appearance data. The comparison of the LSTM based model on HUST-LEBW [9] against state-of-the-art techniques showed how effective it was at detecting eye blinks in the wild and how real-time it could operate.

Most recently, to determine whether or not there is a blink or multiple blinks exist in a particular image sequence, Kajo *et al.* [6] proposed different supervised and unsupervised learning approaches to provide an effective and robust eye blinking detection framework, a using an end-to-end 2D and 3D lightweight CNNs called Pyramidal Bottleneck Block Networks (PBBN). To detect numerous eye blinks, the authors recommended combining moving windowed-Singular Value Decomposition (SVD) with 2D PBBN. In this study, 3D PBBN was primarily employed to identify a single blink in a picture sequence. However, the authors interpolated input images to the same size 96×96 with no further explanation in the preprocessing step, and their best performing model with three pyramids and three branches (P3B3) remained having a high number of parameters, leading to higher inference time on new sequences.

Therefore, in this paper, further evaluation of different interpolation methods to a target image resolution is provided, and the number of model parameters is continued to be reduced in attempt to achieve real time performance, while remaining having sufficient feature extraction ability.

## III. PREPROCESSING OF EYE IMAGES AND DEPTH-WISE SEPARABLE CONVOLUTION FOR A 3D SPATIO-TEMPORAL CNN

In this section, a lightweight Depth-wise Separable Convolution Module for 3D spatiotemporal CNN with Pyramid Bottleneck Block Network (DWS-3D-PBBN) is proposed to determine eye blinking within an image sequence of eye states. The model is built on top of the recent 3D Pyramidal convolution neural network (3D-PBBN) [6] to further shrink the model parameters and reduce inference time for real-time application. Firstly, a general pipeline for eye blinking detection is illustrated in Fig. 1. Then, the paper primarily focuses on the step of preprocessing to address the issue of extracted eye images with multi resolutions, and the step of eye blinking detection model to present architecture of the proposed DWS-3D-PBBN, compared to the baseline model 3D-PBBN.

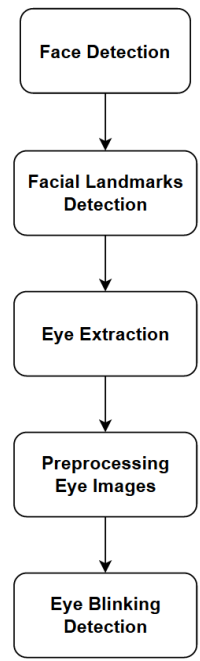

Figure 1. Flow-chart of eye blinking detection.

### A. Workflow of Eye Blinking Detection

Upon receiving a new image in real-time scenario, the workflow for eye blinking detection approach begins with face detection and eye extraction, similar to the workflow proposed in [6]. Here, the Single Shot Detector (SSD) [14] framework based on a ResNet model can be used to detect the face in an image. Once the face is detected, facial landmarks are detected using the fast Kazemi algorithm [15] to identify 68 facial coordinates surrounding certain parts of the face such as the nose, mouth, and eyes. Next, the right and left eye images are extracted by cropping from the landmark points (37 to 42) and (43 to 48) respectively, with a padding of 25% from all directions. The extracted eye images of an incoming frame are then preprocessed and used as input for the eye blinking detection model.

### B. Preprocessing of Eye Images

This section dives further into the evaluation of preprocessing methods relating to image resolutions, and derives optimal ones for suitable tasks, which hasn't been mentioned in previous work. Since the eye images are extracted from scenes under different scenarios, i.e, varying distances from the eyes to camera, the resulting cropped eye images have different resolutions. The preprocessing step ensures that the eye images are of consistent size and format, making them ready to be fed into the detection model. This step involves upsampling or downsampling the image to a target size. For a new image to be processed, if its size is smaller than a target size, the upsampling method is utilized to generate more features for the image, otherwise, downsampling is applied instead, to keep the input size persistent for the model, without significant loss in features. This can be realized by choosing a suitable target size. In our experiment, the mean size of all cropped eye images from the training dataset is selected as the suitable target size, in order to avoid over-upsampling or over-downsampling, resulting in inaccurate features or missing important features. Five interpolation methods (NN, BL, Area, BC, Lanczos) are evaluated in this paper to select the suitable one for upsampling and downsampling to the target size. The resized image is then normalized to facilitate training of the detection model.

### C. Eye Blinking Detection Model

*1) Baseline model: 3D spatiotemporal CNN with Pyramid Bottleneck Block Network (3D-PBBN)*

Inspired by the success of bottleneck residual block [6], authors of the baseline model [6] proposed a simple Pyramid Bottleneck (PB) block that can be utilized for both 2D and 3D inputs. The main motivation of a PB block is to learn multi-resolution features from an input, while reducing the total number of blocks in a network architecture, leading to fewer network parameters. A small number of parameters is essential for achieving real-time performance by shortening the inference time, and to adapt the model's size to the size of the training set, as the eye blink dataset is not substantial enough to train a model with a high number of parameters.

For eye blinking detection from a sequence of frames, the input data is interpreted as a 3D tensor, with two spatial dimensions and one temporal dimension, and the 3D PB block is used to learn the spatiotemporal features. An overview of a 3D PBBN is given in Fig. 2. The model is composed of a starting block of conventional 3D CNN with depth information, multiple 3D PB blocks, and an output block with global average pooling connected with a fully connected layer with two outputs followed by a softmax layer.

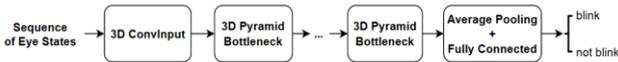

Figure 2. General architecture of a 3D PBBN.

The starting block contains 3D convolutional layer that filters a $D_F \times D_F \times D_T \times 3$ input sequence with K kernels of size 3×3×3 and stride 1×1×2 to downsample the temporal dimension, where $D_F$ is image height/width, and $D_T$ is the number of frames. The convolutional layer is followed by a batch normalization layer, ReLU layer and a 3×3×3 max pooling layer with stride 1×1×2 that continues to downsample by half the temporal dimension, while retaining spatial and channel dimensions.

Then, the starting block is followed by multiple 3D PB blocks downsampling the spatial dimensions to half and doubling channel dimension after each PB block. Every 3D PB block shaped like a pyramid contains several branches, within which consists of multiple layers, and the number of layers corresponds to the current branch number. Assume the current branch is $l^{th}$, it starts with a convolution layer with filter size of $(2l-1) \times (2l-1)$, and keeps reducing by halves in the next convolutional layer within the same branch. An example of a 3D PB block with three branches is illustrated in Fig. 3.

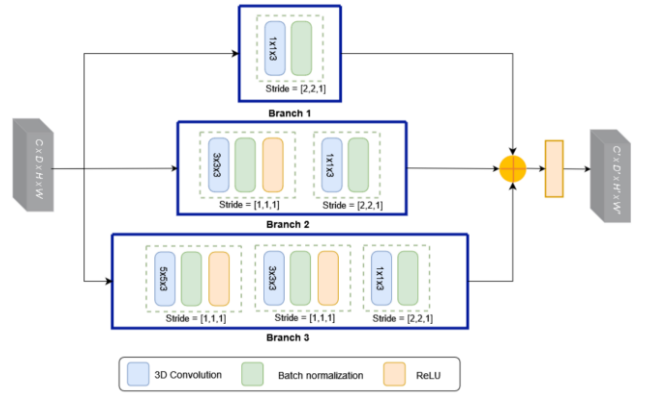

Figure 3. Example of a 3D PB block with 3 branches.

It can be seen from Table I that the number of parameters increases rapidly along with the total number of branches. However, the more branches the PB block contains, the more resolutions of features can be learned through multi-resolution convolution layers at the start of each branch, and thus it is expected to outperform PB blocks with fewer branches in the feature extraction ability. Therefore, it is essential to reduce the number of network parameters while increasing the number of branches without significant loss of feature extraction ability, and one potential approach can be the use of depth-wise separable convolution replacing conventional convolution layers within each branch.

TABLE I. NUMBER OF PARAMETERS FOR INCREASING NUMBER OF BRANCHES IN TWO-PYRAMID AND THREE-PYRAMID NETWORK

| Network | Parameters |
|---------|------------|
| 3D P2B2 | 437442 |
| 3D P2B3 | 1975170 |
| 3D P2B4 | 5934210 |
| 3D P3B2 | 1619650 |
| 3D P3B3 | 7583362 |
| 3D P3B4 | 23243650 |

*2) Proposed model: utilizing depth-wise separable convolution*

   *a) Depth-wise separable convolution*

This section briefly introduces how standard convolution can be factorized into depth-wise convolution and point-wise $1 \times 1$ convolution. A given input feature F with shape $D_F \times D_F \times C$ goes through a standard convolution with kernel $K$ of shape $D_K \times D_K \times C \times C'$, resulting in an output feature map F' shaped $D_{F'} \times D_{F'} \times C'$, where $D_F$ is spatial dimension, $D_K$ is spatial size of kernel, C and C' are the number of input, output channels respectively. The standard convolution (one filter for all input channels) is illustrated in Fig. 4, and can be formalized as $F' = F * K$.

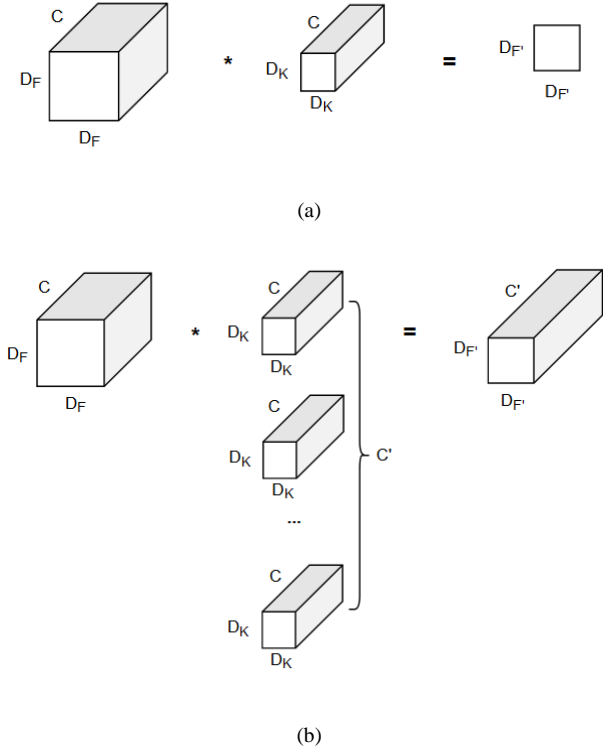

Figure 4. Standard convolution layer. (a) 1 filter with C channels, convolves with C input channels, resulting in 1 output channel. (b) C' filters with C channels, convolves with C input channels, resulting in C' output channels.

In order to drastically reduce the model parameters and computations in standard convolution, depth-wise separable convolution splits the original operation into filter operation - depth-wise convolution (one filter for one input channel), and combination operation - point-wise 1×1 convolution (linear combination of the output depth-wise layer). The filter operation is shown in Fig. 5a, and can be formalized as $\hat{F} = F * K_{dw}$, in which $K_{dw}$ has the shape of $D_K \times D_K \times C$, resulting in $\hat{F}$ shaped $D_{F'} \times D_{F'} \times C'$. It should be noted that F and $\hat{F}$ share the same number of channels. The combination operation is shown in Fig. 5b, and can be formalized as $F' = \hat{F} * K_{dw}$, in which $K_{pw}$ has the shape of $1 \times 1 \times C \times C'$, resulting in F' shaped $D_{F'} \times D_{F'} \times C'$.

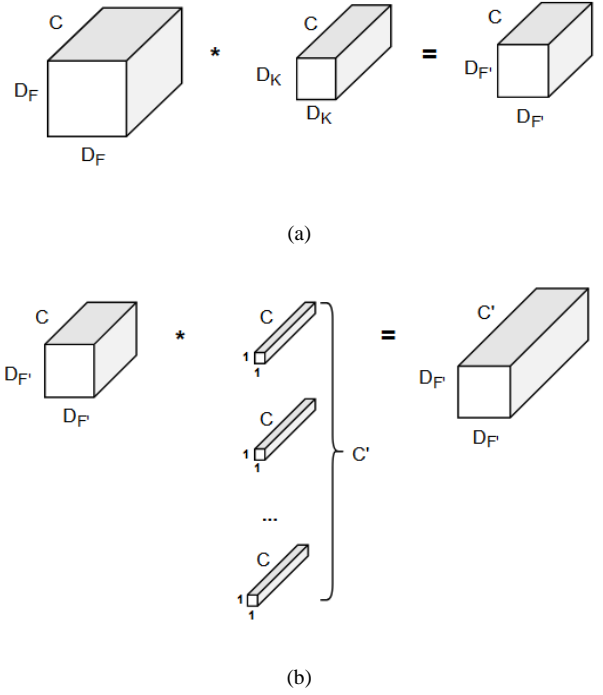

Figure 5. Depth-wise separable convolution. (a) C filters with q1 channel, convolves with C input channels (depth-wise), resulting in C output channels. (b) C' 1×1 filter with C channels, convolves with C previous output channels (point-wise), resulting in C' output channels.

The number of parameters to be optimized in standard convolution is $D_K \times D_K \times C \times C'$. In contrast, depth-wise separable convolution optimizes parameters of lower number $D_K \times D_K \times C + C \times C'$, leading to a saving ratio of $\frac{D_K \times D_K \times C + C \times C'}{D_K \times D_K \times C \times C'} = \frac{1}{C'} + \frac{1}{D_K^2}$. The number of output channels C' and kernel size $D_K$ are generally large in real world scenarios ($C' \gg 1$, $D_K^2 \gg 1$) [11], and thus depth-wise separable convolution can substantially shrink the number of parameters and computations, which is suitable for real-time application.

   *b) Depth-wise separable convolution module for 3D-PBBN*

To improve existing 3D-PBBN, we extended the implementation of depth-wise separable convolution from 2D to 3D convolution layer. While each input channel of 2D convolution contains only two spatial dimensions, the 3D convolution also contains an additional depth dimension. When using depth-wise separable convolution, 2D depth-wise operation applies to two spatial dimensions per input channel, whereas 3D depth-wise operation applies to two spatial dimensions and one depth dimension per input channel. For 3D PBBN, 3D depth-wise separable convolution replaces each 3D convolution layer per branch in each PB block. An example of 3D depth-wise separable Pyramid Bottleneck Block Network with 2 PB block and 2 branches per block (DWS-3D-P2B2) is shown in Table III, replacing the original 3D-P2B2 network in Table II.

TABLE II. ARCHITECTURE OF 3D PBBN THAT CONTAINS TWO PYRAMIDS WITH TWO BRANCHES (3D-P2B2)

| Block | Layer | Filters shape | Stride Size | Output |
|---|---|---|---|---|
| Input | 3DConv | 3×3×3×3×64 | 1×1×2 | 96×96×7×64 |
| | BN | - | - | 96×96×7×64 |
| | ReLU | - | - | 96×96×7×64 |
| | MaxPool | 3×3×3 | 1×1×2 | 96×96×7×64 |
| P1-B1 | 3DConv | 1×1×3×64×64 | 2×2×1 | 48×48×4×64 |
| | BN | - | - | 48×48×4×64 |
| P1-B2 | 3DConv | 3×3×3×64×64 | 1×1×1 | 96×96×4×64 |
| | BN | - | - | 96×96×4×64 |
| | ReLU | - | - | 96×96×4×64 |
| | 3DConv | 1×1×3×64×64 | 2×2×1 | 48×48×4×64 |
| | BN | - | - | 48×48×4×64 |
| Add | ADD | - | - | 48×48×4×64 |
| | ReLU | - | - | 48×48×4×64 |
| P2-B1 | 3DConv | 1×1×3×64×128 | 2×2×1 | 24×24×4×128 |
| | BN | - | - | 24×24×4×128 |
| P2-B2 | 3DConv | 3×3×3×64×128 | 1×1×1 | 48×48×4×128 |
| | BN | - | - | 48×48×4×128 |
| | ReLU | - | - | 48×48×4×128 |
| | 3DConv | 1×1×3×128×128 | 2×2×1 | 24×24×4×128 |
| | BN | - | - | 24×24×4×128 |
| Add | ADD | - | - | 24×24×4×128 |
| | ReLU | - | - | 24×24×4×128 |
| Output | AvgPool | 24×24×4 | - | 1×1×1×128 |
| | FC | 1×128 | - | 2 |

TABLE III. ARCHITECTURE OF DEPTH-WISE SEPARABLE 3D PBBN THAT CONTAINS 2 PYRAMIDS WITH 2 BRANCHES (DWS-3D-P2B2)

| Block | Layer | Filters shape | Stride Size | Output |
|---|---|---|---|---|
| Input | 3DConv | 3×3×3×3×64 | 1×1×2 | 96×96×7×64 |
| | BN | - | - | 96×96×7×64 |
| | ReLU | - | - | 96×96×7×64 |
| | MaxPool | 3×3×3 | 1×1×2 | 96×96×7×64 |
| P1-B1 | 3DConv | 1×1×3×64×64 | 2×2×1 | 48×48×4×64 |
| | BN | - | - | 48×48×4×64 |
| P1-B2 | 3DConv | 3×3×3×64×64 | 1×1×1 | 96×96×7×64 |
| | BN | - | - | 96×96×7×64 |
| | ReLU | - | - | 96×96×7×64 |
| | 3DConv | 1×1×3×64×64 | 2×2×1 | 48×48×4×64 |
| | BN | - | - | 48×48×4×64 |
| Add | ADD | - | - | 48×48×4×64 |
| | ReLU | - | - | 48×48×4×64 |
| Output | AvgPool | 48×48×4 | - | 1×1×1×64 |
| | FC | 1×64 | - | 2 |

## IV. EXPERIMENTS AND RESULTS

### A. Dataset

In order to evaluate eye blinking detection in the wild, the dataset should be chosen to tackle challenges in uncontrolled conditions, and illumination changes sensitivity. Therefore, the dataset HUST-LEBW [9] is selected in this work, which originated from 20 movies and series. Each video has a resolution of 1456×600 or 1200×720, depicting characters under different poses, viewpoint positions and lightning conditions. The videos were split into multiple sequences labeled either as eye blinking or non- eye blinking scenes. Each sequence contains both 10-frame and 13-frame versions, and the models in this paper were trained on sequences of 13 frames. Table IV describes the distribution of labels for left and right eye images in both training and testing sets.

TABLE IV. LABEL DISTRIBUTION PER EYE IN HUST-LEBW [9] DATASET

| Eye | Blinking | Train | Test |
|---|---|---|---|
| Right | Yes | 256 | 126 |
| | No | 190 | 98 |
| Left | Yes | 243 | 122 |
| | No | 181 | 98 |

### B. Interpolation Effects on Eye Blinking Detection Model When Upsampling and Downsampling

Fig. 6 describes the distribution of training image sizes from the dataset HUST-LEBW [9]. Even though the majority of images have size around 50×50, the experiments selected the mean size 96×96 as the target size for interpolation. This can avoid over-upsampling or over-downsampling, which might lead to low performance in new dataset, due to inaccurate features or missing important features. And thus, images with size smaller than 96×96 are upsampled using five interpolation methods Nearest Neighbor (NN), Bilinear (BL), Area, Bicubic (BC), Lanczos4, and the ones larger than 96×96 are downsampled instead.

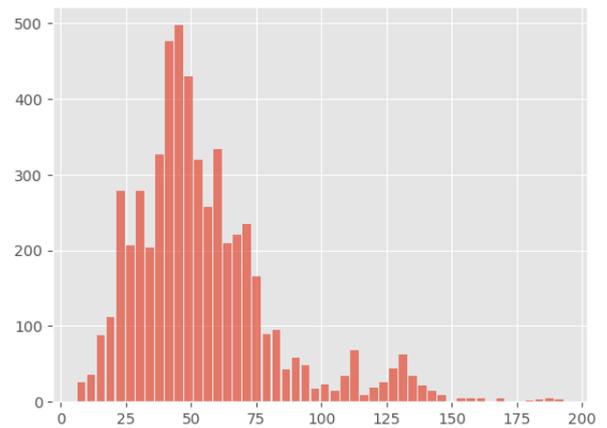

Figure 6. Distribution of image sizes (horizontal axis) over number of images (vertical axis) in the training set of HUST-LEBW [9].

The evaluation metric of each method after training with the proposed model DWS-3D-PBBN 2 pyramids and 2 branches (DWS-3D-P2B2) is done using Precision, Recall, F1-score, which are computed as follows:

$$Recall = \frac{TP}{TP + FN}$$

$$Precision = \frac{TP}{TP + FP}$$

$$F_1 = \frac{2}{\frac{1}{recall} + \frac{1}{precision}}$$

in which TP, TN, FP, FN represent the number of True Positives, True Negatives, False Positives, and False Negatives respectively.

The training runs through a combination of Early Stopping and k-fold cross validations with k=15. The testing sets are extracted according to the type of interpolation, i.e, during evaluation of upsampling, the testing set contains only images smaller than 96×96 from the original testing set. Table V shows the average performance of interpolation methods on a corresponding testing set of only the left eye, with standard deviation after 15 folds. It can be seen that BC obtained both the highest Recall 69.71% and F1-score 71.29% for the purpose of upsampling, whereas BL outperformed others in both Precision 76.77% and F1-score 77.06%. Therefore, BC and BL were selected as preprocessing methods for further evaluation of both original 3D-PBBN and the proposed DWS-3D-PBBN.

TABLE V. AVERAGE PERFORMANCE OF EACH INTERPOLATION METHOD FOR UPSAMPLING AND DOWNSAMPLING AFTER 15-FOLD CROSS-VALIDATION

| Method | Precision | Recall | F1 |
|---|---|---|---|
| UPSAMPLING | | | |
| NN | 0.7559 ± 0.0505 | 0.6717 ± 0.0971 | 0.7068 ± 0.0594 |
| BL | 0.7284 ± 0.0670 | 0.6737 ± 0.1467 | 0.6933 ± 0.1036 |
| Area | 0.7272 ± 0.0672 | 0.6895 ± 0.0819 | 0.7045 ± 0.0558 |
| BC | 0.7382 ± 0.0524 | 0.6971 ± 0.1113 | 0.7129 ± 0.0769 |
| Lanczos4 | 0.7191 ± 0.0471 | 0.6819 ± 0.1474 | 0.6900 ± 0.0949 |
| DOWNSAMPLING | | | |
| NN | 0.6548 ± 0.1241 | 0.8593 ± 0.1755 | 0.7339 ± 0.1224 |
| BL | 0.7677 ± 0.1724 | 0.8222 ± 0.1560 | 0.7706 ± 0.0828 |
| Area | 0.6349 ± 0.0594 | 0.8741 ± 0.1384 | 0.7281 ± 0.0599 |
| BC | 0.6926 ± 0.1476 | 0.8370 ± 0.1319 | 0.7458 ± 0.0934 |
| Lanczos4 | 0.6669 ± 0.1219 | 0.8518 ± 0.1434 | 0.7374 ± 0.0856 |

*C. 3D Pyramid Bottleneck Block Network with Depth-Wise Separable Convolution*

*1) Evaluation of different variants of 3D pyramid bottleneck block network*

Table VI provides a comparison between a deeper (more pyramids) and wider (more branches) of each the original 3D-PBBN and the proposed depth-wise separable convolution approach DWS-3D-PBBN, on left and right eye sequences of HUST-LEBW [9] dataset. Specifically, for each approach, the simple network was implemented with two pyramids and two branches per pyramid, and the complex network was implemented with three pyramids and three branches per pyramid. The experiments were run with 15-fold cross validations, and early stopping with patience of 50 epochs.

Regarding the feature extraction ability for each approach, it can be seen that P3B3 outperforms P2B2 over 2.94% and 1.18%, and DWS-P3B3 outperforms DWS-P2B2 over 0.07% and 0.35% in F1-score for left and right eye images respectively.

Regarding the model size in parameters, it can be seen that the number of parameters of the proposed DWS-P2B2 and DWS-P3B3 are only 13.3% and 6% compared to P2B2 and P3B3 correspondingly. A significant reduction in model size traded off with only 1.63% and 0.3% reduction in Precision performance of DWS-P3B3 compared to P3B3 of each eye. Nevertheless, the Recall scores maintained to increase by 0.82% and 2.01% respectively.

TABLE VI. PERFORMANCE OF DIFFERENT VARIATIONS BETWEEN THE ORIGINAL 3D-PBBN AND DEPTH-WISE SEPARABLE CONVOLUTION DWS-3D-PBBN IN HUST-LEBW [9] DATASET

| Network | | P2B2 | P3B3 | DWS-P2B2 | DWS-P3B3 |
|---|---|---|---|---|---|
| Parameters | | 437442 | 7583362 | 58178 | 455170 |
| Left Eye | Pre | 0.7208 ± 0.0660 | **0.7541 ± 0.0491** | 0.7421 ± 0.0550 | 0.7378 ± 0.0392 |
| | Rec | 0.7207 ± 0.1332 | 0.7360 ± 0.0769 | 0.7415 ± 0.0996 | **0.7442 ± 0.0610** |
| | F1 | 0.7120 ± 0.0803 | **0.7414 ± 0.0375** | 0.7384 ± 0.0630 | 0.7391 ± 0.0323 |
| Right Eye | Pre | 0.7200 ± 0.0420 | **0.7299 ± 0.0435** | 0.7187 ± 0.0504 | 0.7266 ± 0.0709 |
| | Rec | 0.7467 ± 0.0867 | 0.7587 ± 0.0832 | 0.7884 ± 0.0578 | **0.7788 ± 0.0989** |
| | F1 | 0.7286 ± 0.0364 | 0.7404 ± 0.0401 | 0.7433 ± 0.0159 | **0.7468 ± 0.0603** |

Fig. 7 shows the learning curves of different model variants after training with batch size 16, learning rate 0.001, 200 iterations. DWS-P2B2 and DWS-P3B3 have shown to converge to the same loss performance faster than P2B2 and P3B3, whereas P3B3 might need more iterations to gain stability

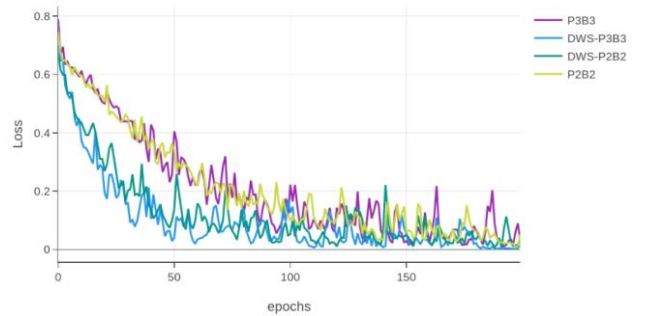

Figure 7. Learning curves of the variants P2B2, P3B3, DWS-P2B2, DWS-P3B3 after 200 epochs, batch size = 16, learning rate = 0.001.

*2) Comparison with baseline models*

Table VII compares the result of the proposed model DWS-P3B3 trained in fold with highest F1-score on the testing set HUST-LEBW [9], with best result of the original P3B3, along with recent methods for eye blinking detection. The table indicates that the KLT tracker based technique proposed by Drutarovsky *et al*. [10] obtained very poor results, since its development was limited to indoor videos and a limited number of persons. The VGG-16 based model proposed by Daza *et al*. [8] obtained the highest Recall score 96.03% for only left eye images, but worse Precision and F1 than other more recent approaches, which is probably due to the constrained database it was trained on, and the inability to capture temporal features. Hu *et al*. [9] with a LSTM based model achieved the highest Recall score 83.33% for right eye images, while Bekhouche *et al*. [6] with 3D-PBBN outperformed other methods in both Precision and F1 of 76.92% and 79.36% respectively in left eye images.

It can be seen that Hu *et al*. [9] and Bekhouche *et al*. [6] are the two methods that leverage both spatio-temporal features for the detection of eye blinking, and they obtained much higher performance in comparison to the previous methods with sole consideration to spatial features. A main advantage of the 3D-PBBN proposed by Bekhouche *et al.* [6] is utilizing pyramid blocks, which enable the ability to extract spatio-temporal features at multi resolutions, without the need for a deeper network.

Meanwhile, the proposed model 3D-PBBN with depth-wise separable convolution (DWS-P3B3) achieved good Precision 82.46% and F1 80.97% scores for sequences of right eye, though performed worse than P3B3 in left eye sequences.

TABLE VII. PERFORMANCE COMPARISON BETWEEN DIFFERENT EYE-BLINKING DETECTION METHODS ON HUST-LEBW [9] DATASET

| Method | Eye Index | Precision | Recall | F1 |
|---|---|---|---|---|
| Drutarovsky *et al*. (2014) [10] | Left | 0.4757 | 0.1190 | 0.1904 |
| | Right | 0.2860 | 0.0952 | 0.1428 |
| Daza *et al*. (2020) [8] | Left | 0.6080 | 0.9603 | 0.7446 |
| | Right | 0.7348 | 0.7950 | 0.7637 |
| Hu *et al*. (2020) [9] | Left | 0.7385 | 0.7805 | 0.7589 |
| | Right | 0.7778 | 0.8333 | 0.8046 |
| Bekhouche *et al*. (2022) [6] (P3B3) | Left | 0.7692 | 0.8196 | 0.7936 |
| | Right | 0.7703 | 0.8253 | 0.7969 |
| Proposed DWS-P3B3 | Left | 0.7285 | 0.8360 | 0.7786 |
| | Right | 0.8264 | 0.7936 | 0.8097 |

## V. DISCUSSION

The results of this study in the preprocessing part have shown that the classification performance of the 3D PBBN based model can be considerably affected by different interpolation methods. Bicubic has shown its superior performance in both Recall and F1 of the detection model for the upsampling task, as previously reported in other studies analyzing the effects of interpolation methods on deep learning models [7]. Meanwhile, bilinear method brought the best Precision and F1 of the classifier for the downsampling task, which is different to the Area method reported by OpenCV for being usually favorable in practice for image decimation [12].

Importantly, regarding the eye blinking detection models using PBBN-based approach, increasing the number of pyramids and the number of branches per pyramid still generally showed an increase in its performance for all the evaluated metrics Precision, Recall, and F1, according to Table VII, but trading off with increasing the model size. Therefore, our proposed model DWS-PBBN, implementing depth-wise separable convolution into each convolutional layer within every branch of each pyramid, has significantly reduced the number of parameters when increasing the number of pyramids and branches, allowing to have a more complex network with higher feature extraction ability. The results after 15-fold cross validation have shown that the proposed DWS-P3B3 could not only reduce the original model parameters to only 6%, but also outperformed the original P3B3 in Recall metric of both left and right eye sequences. A model with higher Recall means having lower false negatives, which also means having lower chance of missing an eye-blink. This might be important for some tasks in which the detection of an eye-blink is critical, such as driver drowsiness detection, liveness detection or face anti-spoofing.

However, there remains certain challenges and limitations in the evaluated dataset HUST-LEBW [9], such as not fully closed eye samples due to low sampling, artifacts on eyes such as makeup confusing the classifier, dramatic lightning conditions, all of which might result in lower detection performance of both original P3B3 and DWS-P3B3. Examples of misclassification samples due to these limitations are illustrated in Fig. 8. Additionally, a potential future work is to design a model to detect multiple eyeblinks in a sequence, however, HUST-LEBW [9] contains only single eyeblinks, and thus it might not be suitable for this purpose.

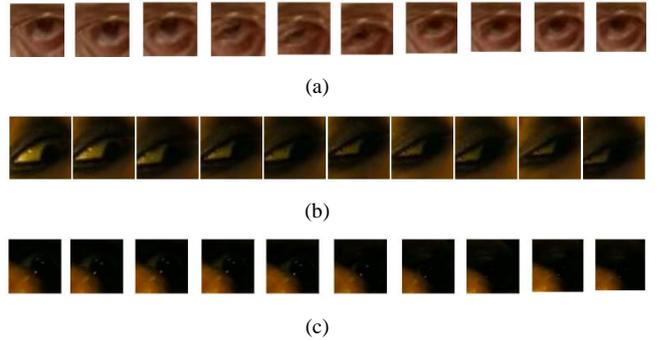

Figure 8. The failure cases of both PBBN and DWS-PBBN approaches evaluating on HUST_LEBW [9]. (a) False negative eyeblink (due to eye not fully closed). (b) False positive non-eyeblink (due to makeup around eye). (c) Dramatic lightning condition (eye features are barely visible in some samples).

## VI. CONCLUSION

In order to improve eye blinking recognition, the findings of this study's preprocessing section have demonstrated that various interpolation techniques may significantly alter the classification performance of the 3D PBBN-based model. For the upsampling task, Bicubic is one of five interpolation algorithms that has demonstrated its superior performance in both Recall and F1 of the detection model. In the meanwhile, the Bilinear method provided the classifier with the best Precision and F1 for the downsampling task. In addition, this paper utilized depth-wise separable convolution to increase model complexity while reducing the number of network parameters. The proposed DWS-PBBN has increased the model complexity from DWS-P2B2 to DWS-P3B3 to obtain higher feature extraction ability, while keeping the number of parameters under 6% with a sacrifice of only 1.63% and 0.33% in the Precision of left and right eye respectively, in comparison with P3B3. Additionally, the proposed model DWS-PBBN also maintained to outperformed PBBN in the Recall metric in most scenarios, which is essential for some eye blinking applications where the missing of true eye blinks pays a high cost. Our source code can be found at: https://github.com/bachzz/DWS-PBBN

As future work, we will attempt to investigate further into real-time constraints and implement two global hyper-parameters width multiplier and resolution multiplier proposed in MobileNets [13], in order to address the tradeoff between latency and accuracy of the model. These hyper-parameters allow the depth-wise separable convolution model to be designed with the right size depending on the resource and accuracy of the problem, which is real time eye blinking detection in this case.

## CONFLICT OF INTEREST

The authors declare no conflict of interest.

## AUTHOR CONTRIBUTIONS

Nguy Thi Lan Anh conceptualized, designed depth-wise convolution layers, and executed the majority of experiments, then contributed to writing the original draft manuscript. Nguyen Gia Bach and Phan Xuan Tan reviewed related studies, supported running experiments, and contributed to writing original draft manuscripts. Phan Xuan Tan, Nguyen Thi Thanh Tu and Eiji Kamioka supervised, then reviewed and edited the manuscript. All authors have read and agreed to the published version of the manuscript.